\RequirePackage{luatex85}

\documentclass{article}
\usepackage{ijcai22}

\usepackage{times}
\usepackage{soul}
\usepackage{url}
\usepackage[hidelinks]{hyperref}
\usepackage[utf8]{inputenc}
\usepackage[small]{caption}
\usepackage{graphicx}
\usepackage{subcaption}
\usepackage[usenames,dvipsnames]{xcolor}
\usepackage{amsmath}
\usepackage{amsthm}
\usepackage{booktabs}
\usepackage{algorithm}
\usepackage{algorithmic}
\usepackage{tikz}
\usetikzlibrary{
	arrows,
	automata,
	positioning,
	shapes,
	calc,
	graphs,
	quotes,
	matrix,
	backgrounds,
	fit,
	external,
	angles}

\tikzset{shownodename/.style={
			append after command={
					\pgfextra{\node [right] at (\tikzlastnode.mid east) {\tikzlastnode};}
				}
		}
}
\tikzstyle{base node}=[draw, align=center, text centered, anchor=center]

\title{Efficient entity-based reinforcement learning}

\author{Vince Jankovics\and
	Michael Garcia Ortiz\and
	Eduardo Alonso
	\affiliations
	City, University of London
	\emails
	\{vince.jankovics, michael.garcia-ortiz, e.alonso\}@city.ac.uk
}

\begin{document}

\maketitle

\begin{abstract}

	Recent deep reinforcement learning (DRL) successes rely on end-to-end learning
	from fixed-size observational inputs (e.g.\ image, state-variables). However,
	many challenging and interesting problems in decision making involve
	observations or intermediary representations which are best described as a set
	of entities: either the image-based approach would miss small but important
	details in the observations (e.g.\ objects on a radar, vehicles on satellite
	images, etc.), the number of sensed objects is not fixed (e.g.\ robotic
	manipulation), or the problem simply cannot be represented in a meaningful way
	as an image (e.g.\ power grid control, or logistics).

	This type of structured representations is not directly compatible with
	current DRL architectures, however, there has been an increase in machine
	learning techniques directly targeting structured information, potentially
	addressing this issue.

	We propose to combine recent advances in set representations with slot
	attention and graph neural networks to process structured data, broadening the
	range of applications of DRL algorithms. This approach allows to address
	entity-based problems in an efficient and scalable way. We show that it can
	improve training time and robustness significantly, and demonstrate their
	potential to handle structured as well as purely visual domains, on multiple
	environments from the Atari Learning Environment and Simple Playgrounds.
\end{abstract}

\section{Introduction}

Reinforcement learning is concerned with how agents can take decisions based on
their observations and act optimally in their environment. Recently, deep
reinforcement learning has incorporated deep neural networks to learn how to act
directly from raw observations, by interacting with the environment. It was
demonstrated on very challenging tasks that it was possible to learn very
complex behaviors (walking, navigation), occasionally surpassing humans (Atari,
Go,
Starcraft)\cite{vinyalsGrandmasterLevelStarCraft2019,badiaAgent57OutperformingAtari2020,schrittwieserMasteringAtariGo2019}.

Most of DRL models learn a policy from raw sensory data, such as image (R,G,B),
joints, or representation of a game as an image (Go). This fixed-size data
format is ubiquitous in the field, because it is appropriate for a large variety
of problems where observations provided to the agents are pre-determined and
always in the same format, and because the fixed memory size allows to leverage
parallel processing to dramatically accelerate learning.

However, this approach is not suitable for all decision making problems. First
of all, certain problems cannot be efficiently represented as fixed-size
tensors. Solving problems that require reasoning on a dynamically changing
observation space (i.e.\ where the closed-world assumption does not hold) has
always been challenging for artificial
intelligence~\cite{milchBLOGRelationalModeling2004}. For example, tasks such as
the control of a power grid or the tracking of objects by nature represent the
observation space as a set of entities. The solution often proposed is to
convert the observation space into a fixed size encoding, which is suboptimal.
That way the structure of the problem is lost and has to be rediscovered by the
DNN architecture during training.

Even in the case of observation spaces that are naturally static size, certain
problems can benefit from converting these observations into structured
representations. For example, a robot operating in a human environment receives
as inputs RGB camera images, but these images can be converted into objects that
constitutes the scene (and their properties). Operating at the level of entities
potentially has many advantages: representations are more compact, disentangled,
and knowledge transfer is better. Compositional reasoning using objects,
relations, and actions is crucial for human
cognition~\cite{spelkeCoreKnowledge2007}, which motivated several recent works
in machine learning to represent scenes as a set of objects and their
properties, and the relations between
them~\cite{battagliaRelationalInductiveBiases2018,changCompositionalObjectbasedApproach2016}.
\begin{figure*}[h]
	\centering
	\begin{subfigure}[b]{0.25\textwidth}
		\centering
		\includegraphics[width=\textwidth]{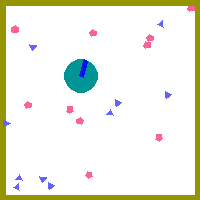}
		\caption{}\label{fig:candy_poison}
	\end{subfigure}%
	~
	\begin{subfigure}[b]{0.25\textwidth}
		\centering
		\includegraphics[width=\textwidth]{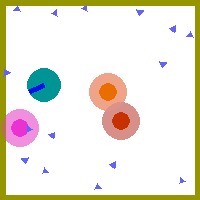}
		\caption{}\label{fig:candy_fireballs}
	\end{subfigure}%
	~
	\begin{subfigure}[b]{0.25\textwidth}
		\centering
		\includegraphics[width=\textwidth]{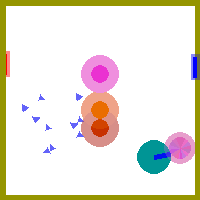}
		\caption{}\label{fig:dispenser_fireballs}
	\end{subfigure}%
	\caption{The three SPG learning scenarios with the agent (cyan circle) and the
	  entities the agent can interact with. (\subref{fig:candy_poison}) $CandyPoison$: the small blue
	  triangles are $Candies$ and the pink pentagons are $Poisons$
	  (\subref{fig:candy_fireballs}) $CandyFireballs$: the pink, orange and red circles are the Fireballs with their transparent interaction radius.
	  (\subref{fig:dispenser_fireballs}) $DispenserFireballs$:
	  The pink circle at the bottom-left corner is a $Dispenser$ that drops $Candies$ at the opposite side of the map, and the read and blue rectangles are two $Portals$ .}\label{fig:scenarios}
\end{figure*}

In this paper, we propose to address this challenge by combining different
neural architectures allowing to process sets and graphs, so that an agent can
learn from structured representations in a seamless, end-to-end fashion.

In order to test our proposed entity-based policy architecture, in the case when
entities are not provided by the environment, we use a simple color-based
segmentation module, since we are dealing with games that have very simple
graphics. However, this approach can be extended with a pre-trained object
detection model that would enable this approach to be used on real-world
problems.

We use a standard actor-critic policy with a shared
encoder~\cite{andrychowiczWhatMattersOnPolicy2020}. We only focus on the choice
of the encoder (while keeping the actor and critic branches the same), and
compare how this encoder performs when considering the observation space as a
set or as a graph. slot-attention (SA)
module~\cite{locatelloObjectCentricLearningSlot2020} is used to process the
observation space as a set, and a graph-attention (GAT)
module~\cite{brodyHowAttentiveAre2021} is used when processing the observation
space as a graph.

Our contributions are as follows:
\begin{enumerate}
	\item We propose an end-to-end architecture that learns from a structured
	      dynamic size representation of the agent's observations. This enables
	      the agents to solve complex tasks in a more sample efficient way.
	\item We provide a set of environments based on Simple Playgrounds
	      (SPG)~\cite{garciaortizSimpleplaygrounds2021}, a light-weight, simple
	      and customisable 2D simulation environment. These environments allow a
	      direct comparison between entity-based RL and more classical image-based
	      RL (see \figurename{~\ref{fig:scenarios}}).
	\item We present a thorough comparison between our approach and standard
	      approaches of the field on these environments, as well as on the Atari
	      Learning Environment (ALE) benchmark.
	\item We provide clean, tested code using PyTorch and
	      RLlib~\cite{liangRLlibAbstractionsDistributed2018} with the required
	      configurations of parameters, which allows easy reproduction of the
	      results, reuse of the model architectures. We hope that this will help
	      further research in this direction.\footnote{All the code used to
		      produce the results and config files are available at
		      \url{anonymous.4open.science/r/spg-experiments/}} \footnote{For videos
		      of the agents performing in the different scenarios please see the
		      playlist on
		      \url{youtube.com/playlist?list=PL0fzH_bs_m9jy4uzf8Oj5TP11OVEPxATh}}
\end{enumerate}

\section{Related work}

Using abstract structures in RL problems has been studied for a long time.
In~\cite{dzeroskiRelationalReinforcementLearning2001} the authors explore the
use of relational learning to induce general rules that an agent can utilise,
while
in~\cite{guestrinGeneralizingPlansNew2003,diukObjectorientedRepresentationEfficient2008}
the authors formulate an object-based Markov Decision Process (MDP) approach
that directly models the environment in terms of objects and their interactions.
These approaches allow to train agents on structured data, howevever they do not
scale to complex problems that require the use of DRL. In more recent works,
~\cite{garneloDeepSymbolicReinforcement2016} explore neuro-symbolic policies,
however it is not clear if their approach would scale beyond their test
environment.

In ~\cite{bakerEmergentToolUse2020}, the authors propose to train agents using
structured observations (set of entities in the scene and their properties).
However, they do not address the problem of varying input size. Each agent has a
fixed amount of slots to represent each entity (one per slot), and these slots
are masked with mask-attention when the corresponding entities are not in the
field of view of the agent. We argue that this approach has several limitations.
First of all, the total possible number of entities needs to be known in advance
by the designer, which is not flexible, and does not allow transfer to new
environments. Then, the model accounts for all entities and masks the ones it is
not supposed to see to focus on only the observed entities, which is wasteful
and significantly less efficient. In real-world environments where a variable
number of entities can be present this approach would not scale.

\begin{figure*}[h]
	\centering
	\begin{subfigure}[b]{0.4\textwidth}
		\centering
		\includegraphics[width=\textwidth]{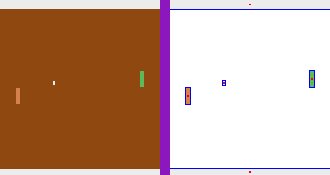}
		\caption{}\label{fig:pong}
	\end{subfigure}%
	~
	\begin{subfigure}[b]{0.4\textwidth}
		\centering
		\includegraphics[width=\textwidth]{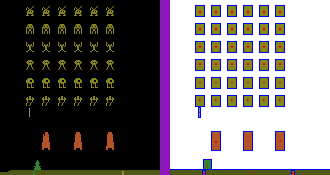}
		\caption{}\label{fig:space_invaders}
	\end{subfigure}%
	\caption{Example Atari games as the raw observation (left) and the extracted entities
	  visualized (right). The features for each entity includes the color or the
	  entity, the location of the centroid (red dot) and the size of the
	  bounding box (blue rectangle). (\subref{fig:pong}) demonstrates this on
	  Pong and (\subref{fig:space_invaders}) on SpaceInvaders.}\label{fig:atari}
\end{figure*}

There has been an increase in algorithms that operate directly on sets, where we
expect the model to be invariant to the order of the input. E.g.\ the
self-attention module~\cite{vaswaniAttentionAllYou2017} is often used, but it is
limited to deal with sets of constant cardinality (i.e.\ number of set
elements). The slot-attention
module~\cite{locatelloObjectCentricLearningSlot2020} can learn representations
from sets of different cardinality. It is important to note that in principle
the size of the input set does not need to be known in advance, but in practice
only fixed-size problems have utilized this algorithm (for computational and
implementation reasons). We utilise the theoretical flexibility of the
slot-attention module and provide an efficient implementation that enables
learning on variable size input spaces.

Graph neural networks (GNN) solve a similar problem, but with the possibility to
capture relationships between entities as edges of the input graph. GNN-based RL
approach has been explored in domain specific problems (
e.g.~\cite{marotLearningRunPower2021,gammelliGraphNeuralNetwork2021}). Even
though it can yield good results if enough resources are
available~\cite{vinyalsGrandmasterLevelStarCraft2019,bakerEmergentToolUse2020},
we we argue that in general an object based approach could outperform approaches
that rely on a fixed length input vector with zero-padding.

There have been attempts to use GNNs to learn a structured policy on various
domains~\cite{wangNerveNetLearningStructured2018,kipfContrastiveLearningStructured2019},
but they only focused on fixed observation spaces, and focus on learning a
structured latent representation, which limit their use to closed-world problems
and make them impractical for real-world applications.

\section{Methods}

In this section we describe first the environments that we used for testing, the
visual and entity-based approaches and the RL policy structure in more details.

\subsection{Environments}

In order to assess and evaluate the proposed methods for entity-based
Reinforcement Learning, we use two simulation environments. The Atari Learning
Environment (ALE) allows us to present results on established DRL benchmarks to
validate the approach. However, Entity-based representations are not directly
accessible in ALE. In order to conduct a thorough and fair comparison between
vision-based and entity-based RL, we use
Simple-Playgrounds~\cite{garciaortizSimpleplaygrounds2021}, that allows both
visual observations (top-down or from the point of view of the agent) and
entity-based observations.

For our experiments on ALE, we use the standard action space without any
modifications. We benchmark our method on Pong, but any game can be processed
with our pipeline, as \figurename{~\ref{fig:atari}} demonstrates. In the
remaining of the section, we will focus more in detail on the environments
developed with the SPG simulator.

\paragraph{Simple Playgrounds}

In SPG, an agent moves in a room and interacts with different entities. To
simplify, we limited the interactions to contact interactions, triggered when an
agent touches or is in the close proximity of an entity.

The actions of the agent in SPG are 2 independent discrete values, the first for
forward and backward motion and the second for rotation around the centre of the
agent, both taking values from $\{-1, 0, 1\}$ where 0 is the no-action.
\begin{figure*}[h]
	\definecolor{c_agent}{RGB}{0,150,150}
	\definecolor{c_candy}{RGB}{0,100,250}
	\definecolor{c_portal_red}{RGB}{255,0,0}
	\definecolor{c_portal_blue}{RGB}{0,0,255}
	\definecolor{c_dispenser}{RGB}{235,50,210}
	\definecolor{c_fireball}{RGB}{235,110,0}
	\tikzstyle{base_node}=[draw, circle, minimum size=10pt, align=center]
	\tikzstyle{agent}=[base_node, fill=c_agent]
	\tikzstyle{candy}=[base_node, fill=c_candy]
	\tikzstyle{portal_red}=[base_node, fill=c_portal_red]
	\tikzstyle{portal_blue}=[base_node, fill=c_portal_blue]
	\tikzstyle{dispenser}=[base_node, fill=c_dispenser]
	\tikzstyle{fireball}=[base_node, fill=c_fireball]
	\tikzstyle{arr}=[->, bend left=7]
	\tikzstyle{no_spring}=[spring constant=1000000]
	\tikzstyle{dir_graph}=[-, >=stealth',shorten >=1pt, shorten <=1pt, auto,
	node distance=1.5cm]
	\centering
	\begin{subfigure}[b]{0.25\textwidth}
		\centering
		\includegraphics[width=\textwidth]{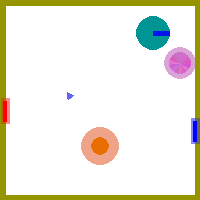}
		\caption{}\label{fig:topdown}
	\end{subfigure}%
	~
	\begin{subfigure}[b]{0.25\textwidth}
		\centering
		\begin{tikzpicture}[dir_graph]

			\def \n {6}
			\def \offset {150}
			\def \radius {1.8cm}

			\node [portal_red] at ({\offset + 360/\n * 0}:\radius) (R) {};
			\node [portal_blue] at ({\offset + 360/\n * 1}:\radius) (B) {};
			\node [dispenser] at ({\offset + 360/\n * 2}:\radius) (D) {};
			\node [agent] at ({\offset + 360/\n * 3}:\radius) (A) {};
			\node [candy] at ({\offset + 360/\n * 4}:\radius) (C) {};
			\node [fireball] at ({\offset + 360/\n * 5}:\radius) (F) {};

			\draw [arr] (R) edge (A) edge (B) edge (C) edge (D) edge (F);
			\draw [arr] (A) edge (B) edge (C) edge (D) edge (F) edge (R);
			\draw [arr] (B) edge (C) edge (D) edge (F) edge (R) edge (A);
			\draw [arr] (C) edge (D) edge (F) edge (R) edge (A) edge (B);
			\draw [arr] (D) edge (F) edge (R) edge (A) edge (B) edge (C);
			\draw [arr] (F) edge (R) edge (A) edge (B) edge (C) edge (D);
		\end{tikzpicture}
		\vspace*{0.5cm}
		\caption{}\label{fig:topdown-graph-full}
	\end{subfigure}%
	~
	\begin{subfigure}[b]{0.25\textwidth}
		\centering
		\begin{tikzpicture}[dir_graph]

			\def \n {6}
			\def \offset {150}
			\def \radius {1.8cm}

			\node [portal_red] at ({\offset + 360/\n * 0}:\radius) (R) {};
			\node [portal_blue] at ({\offset + 360/\n * 1}:\radius) (B) {};
			\node [dispenser] at ({\offset + 360/\n * 2}:\radius) (D) {};
			\node [agent] at ({\offset + 360/\n * 3}:\radius) (A) {};
			\node [candy] at ({\offset + 360/\n * 4}:\radius) (C) {};
			\node [fireball] at ({\offset + 360/\n * 5}:\radius) (F) {};

			\draw [arr] (A) edge (D) edge (C);
			\draw [arr] (D) edge (B) edge (A);
			\draw [arr] (B) edge (D) edge (F);
			\draw [arr] (R) edge (C) edge (F);
			\draw [arr] (C) edge (R) edge (F);
			\draw [arr] (F) edge (R) edge (C);
			\draw [arr, BrickRed] (B) edge (R);
			\draw [arr, BrickRed] (R) edge (B);
		\end{tikzpicture}
		\vspace*{0.5cm}
		\caption{}\label{fig:topdown-graph-2n}
	\end{subfigure}%
	\caption{(\subref{fig:topdown}) a simplified $DispenserFireballs$ scenario,
	  (\subref{fig:topdown-graph-full}) as a fully connected graph and
	  (\subref{fig:topdown-graph-2n}) represented as a graph with connections
	  between the 2 nearest neighbours and a connection between the
	  portals.}\label{fig:playg}
\end{figure*}
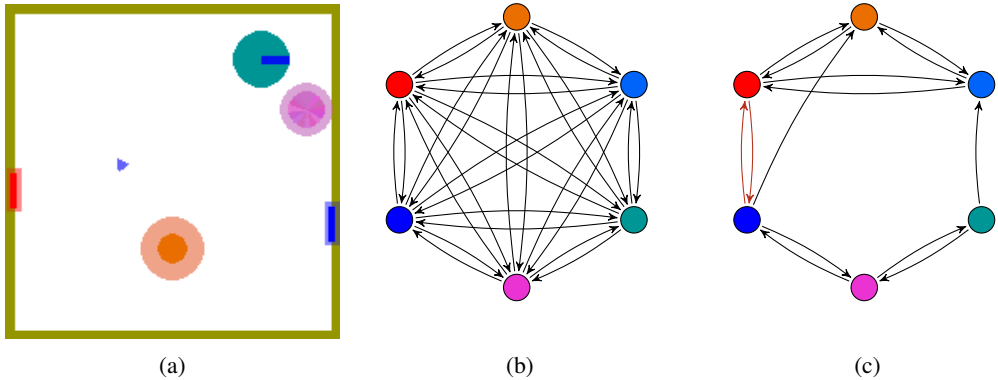

In our proposed scenarios, an agent can encounter different kinds of entities:
\begin{itemize}
	\item $Walls$ are fixed, non movable. They surround the limits of the
	      playground making it a closed environment (room).
	\item $Candies$ are absorbed on contact and provide a reward of +5.
	\item $Poisons$ are absorbed on contact and provide a negative reward of -5.
	\item $Fireballs$ are moving in the environment, following a pre-set
	      trajectory, and provide a negative reward from $\{-5, -2, -1\}$
	      (depending on the color) when in contact with the agent.
	\item $Dispensers$ are fixed and generate Candies in another location of the
	      playground when in contact with the agent.
	\item Red $Portal$ teleports the agent to Blue $Portal$ and vice et versa.
\end{itemize}

These simple entities allow to create RL tasks which require non-trivial
sequential decision making. Using $Portals$ also invalidates the Euclidian space
assumption that agents could usually rely on for navigation tasks.

\paragraph{CandyPoison scenario:} $Poisons$ and $Candies$ are spread randomly in
the playground (see \figurename{~\ref{fig:candy_poison}}). The agent must
collect $Candies$ while avoiding $Poisons$. This scenario is analogous to the
minigame $CollectMineralShards$ from SC2LE~\cite{vinyalsStarCraftIINew2017}.

\paragraph{CandyFireballs scenario:} $Candies$ are spread randomly in the
playground (see \figurename{~\ref{fig:candy_fireballs}}). 3 moving $Fireballs$
that cause negative reward if the agent gets close to them. The agent must learn
to collect $Candies$ while avoiding $Fireballs$. As the trajectory of the
fireball is a sequence of random waypoints, it prevents the agent from
remembering sequences of actions and force it to learn a policy based on its
observations.

\paragraph{DispenserFireballs scenario:} This environment is designed to
demonstrate multi-step planning by including a $Dispenser$ object that needs to
be activated to drop $Candies$ in a random location of the environment (see
\figurename{~\ref{fig:dispenser_fireballs}}). The location change at every
episode. To access the location of the $Candies$, the agent must cross a central
region occupied by a wall of moving $Fireballs$. Alternatively it can use
$Portals$ to avoid this region and teleport directly to the other side of the
playground. The agent needs to learn that there is a shortcut that breaks the
Euclidean symmetry of the space.

The different learning scenarios are illustrated in
\figurename{~\ref{fig:scenarios}} and \figurename{~\ref{fig:atari}}.

\subsection{Visual baselines}

In order to evaluate our entity-based RL approach, we compare it to purely
visual baselines that process raw RGB values with a convolutional neural network
(CNN) encoder as described in~\cite{mnihPlayingAtariDeep2013}.

In ALE, visual observations are provided as RGB images that correspond to a
top-down visualisation of the game. In SPG, this RGB information corresponds to
the agent's view of the surrounding scene as a 1D strip together with the
distance of the observed entities.

\subsection{Entity-based observations}

\paragraph{In SPG:} the simulator allows direct access to the entities present
in the scene, in the form of a sensor that detects non-occluded entities within
a determined range and field of view, In our experiments, we set the range to
400 pixels and the field of view to 360 degrees, and we use the same settings to
construct the entity-based observations, i.e. the amount of information is the
same in both settings.

The observations are ego-centric to the agent, so each object is described by a
feature vector $x = [t, d, \sin(\alpha), \cos(\alpha)]$, where $t$ is a one-hot
encoding of the type of the object (as the number of types are known for each
scenario), $d$ is the distance from the agent and $\alpha$ is the relative angle
of the observed entity. To keep the periodic prior of the angles we use a
$\sin/\cos$ transformation. We normalize $d$ with the maximum observation range.
\begin{figure*}[h]
	\centering
	\begin{subfigure}[b]{0.33\textwidth}
		\centering
		\includegraphics[width=\textwidth]{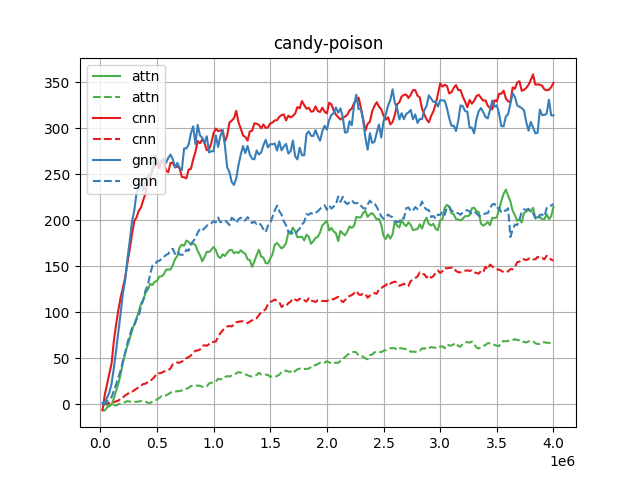}
		\caption{}\label{fig:candy_poison_training}
	\end{subfigure}%
	~
	\begin{subfigure}[b]{0.33\textwidth}
		\centering
		\includegraphics[width=\textwidth]{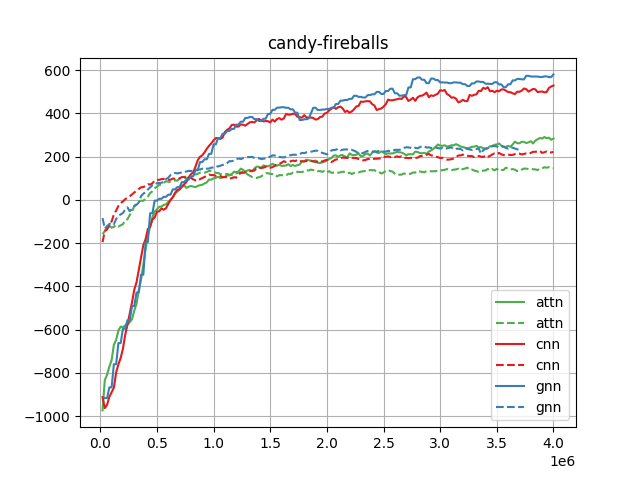}
		\caption{}\label{fig:candy_fireballs_training}
	\end{subfigure}%
	~
	\begin{subfigure}[b]{0.33\textwidth}
		\centering
		\includegraphics[width=\textwidth]{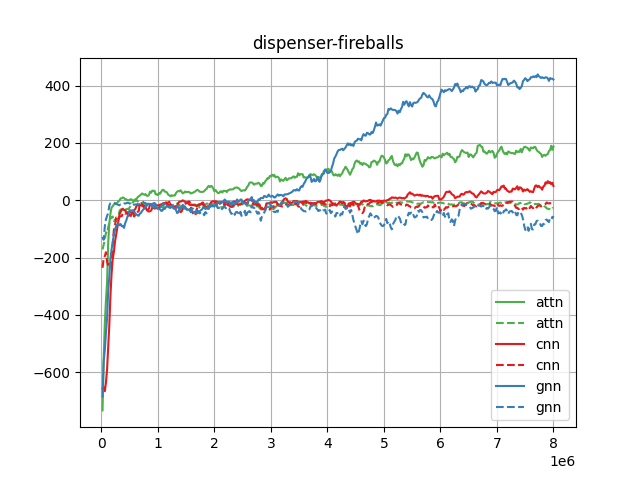}
		\caption{}\label{fig:dispenser_fireballs_training}
	\end{subfigure}%
	\caption{Training curves for all the SPG test scenarios. The solid lines show
		the base scenarios while the dashed are their $large$ versions.
		(\subref{fig:candy_poison_training}) shows $CandyPoison$,
		(\subref{fig:candy_fireballs_training}) $CandyFireballs$ and
		(\subref{fig:dispenser_fireballs_training})
		$DispenserFireballs$.}\label{fig:spg_training}
\end{figure*}

\paragraph{In ALE:} there are no built-in entity-based observations, so we
utilize a pipeline that allows our entity-based policy to be applied to any
purely visual problem. We obtain entities by applying a simple segmentation
algorithm (since entities in the ALE environments we use have a single color).
This algorithm separates the scenes by identifying blobs of regions that have
the same pixel values, and calculates their centroid and bounding box.

Each extracted entity is described by $x = [r, g, b, x, y, dx, dy, s]$, where
$r, g, b$ is the color of the entity, $x, y$ are the absolute coordinates,
$dx, dy$ are bounding box dimensions and $s$ is the stack location of the
feature in the frame-stack. We normalize the coordinates to the height and width
of the game, and $s$ to the number of stacks. Since in general the type of the
entity is not known (unlike in SPG) we rely on a more flexible approach that
uses the color itself. The assumption is that similar entities have similar
colors, which typically holds in Atari.

In more complex visual scenarios we could rely on already existing
high-performance object detection models
(e.g.~\cite{renFasterRCNNRealTime2015}), which would allow our proposed policy
architecture to be used on real-world problems.

The entity-based observations is presented to the policy network as either a set
of entities, or as a graph, where we can capture relationships between the
entities. In this paper we create a fully connected graph from the observed
entities, but arbitrary relationships can be captured with different edge types,
e.g.\ spacial distance, as illustrated in \figurename{~\ref{fig:playg}}.

\subsection{Policies}

\paragraph{Image-based RL:}
For the purely visual case we use a simple CNN to process the observations of
the agent. The difference between SPG and Atari is that in SPG the visual
observations are provided as a 1D strip as the agent perceives its environment,
while in Atari it is a 2D top-down representation. In the SPG case there is also
occlusion, since objects can be in front of each other, while Atari provides
full observability.

We use a matching CNN structure, i.e.\ it is a 1D convolutional encoder for SPG
and 2D for Atari. In both cases the input is projected to a fixed size latent
representation that the actor-critic branches utilise downstream.

\paragraph{Entity-based RL:}

For the entity-based approach we use slot-attention to process entities in the
form of a set, or GNN encoders to process entities in the form of a graph.

The GNN module is based on the graph-attention architecture
(GAT)~\cite{brodyHowAttentiveAre2021} with an additional global mean pool to
create a fixed size output. The GAT with global pooling and slot-attention
modules are fundamentally similar, except that the slot-attention does not pass
information between the input entities (i.e.\ slots), while the GAT module does.
This provides a potentially more expressive architecture.

\paragraph{Pre-processing of the observations:}
For SPG environments, we do not apply any pre-processing to the observations.
For Atari we follow the same pre-processing as described
in~\cite{mnihPlayingAtariDeep2013}, namely frame-skip and frame-stack. While for
the CNN architecture the stacking can be done along the color dimension, in the
entity-based representation we include the stack-depth as part of the input
features, forming a set of entities that captures their temporal evolution. Note
that we do not associate the entities explicitly across frames, i.e.\ it is up
to the policy network to learn the corresponding temporal relationships. This
could be improved with an approach that matches entities between frames, e.g.\
see~\cite{creswellUnsupervisedObjectBasedTransition2021}.

\subsection{RL algorithm}

We use PPO~\cite{schulmanProximalPolicyOptimization2017} as the learning
algorithm that optimises the agent's behaviour in an on-policy setting.

In an actor-critic approach the learnt policy $\pi(a_{t}|s_{t})$ represents the
probability of choosing an action $a_{t}$ given the state $s_{t}$, and
$v_{f}(s_{t})$ provides an estimate of the value function for a given state. We
use a shared latent-space with separate policy and value-function
heads~\cite{andrychowiczWhatMattersOnPolicy2020}.

\section{Experiments}

We compare the CNN, SA and GNN encoder architectures across different scenarios.
The SPG experiments have 3 levels of complexity in the following order
$CandyPoison$, $CandyFireballs$ and $DispenserFireballs$. The final scenario
requires complex spatial reasoning and multi-step planning, since the agent
needs to first touch the $dispenser$ to get the $candy$ dropped at a different
part of the scene.

We also have a large version of each SPG environment, that tests how the
different representations scale with the size of the scene. I.e.\ we keep the
size of the entities the same, but increase the whole environment from 200x200
pixels to 1000x1000. This is similar to the problem where an agent has a large
observation field (e.g.\ radar station, or satellite) with relevant objects
scattered sparsely in the field. We argue that an object-based representation is
invariant to this sparseness. However, it is important to note that the learning
speed (in terms of number of samples) will decrease, because the agent is
required to explore more to learn the optimal behaviour. This issue could be
addressed separately by utilising a curiosity-based exploration algorithm,
e.g.~\cite{camperoLearningAMIGoAdversarially2021}.

The results are detailed in Table \ref{tab:results}. They demonstrate that while
on the simpler environments the CNN and GNN approaches perform similarly, on the
most complex environment the GNN outperforms the CNN by a large margin. Also, in
the scaled-up scenarios the difference between CNN and GNN architectures are
further highlighted. We note, that the SA experiments under perform in most
scenarios the others, probably due to the limited network capacity. We have not
established a thorough lower bound to the required number of parameters for
each, but we managed to achieve comparable performance with a GNN that has ~2.5k
parameters against a CNN that has ~140k. Future work would include an ablation
study to see the more accurately how the size of the encoder impacts
performance.

The results on Pong show a significant increase in the sample efficiency. The
CNN architecture has a flat period until ~2M samples, while the GNN takes off
after ~500k. We note that there is no additional heuristics added to the entity
extraction process, it only relies on the colors of the objects, therefore it
can be applied trivially to other Atari games. However, one current limitation
is that the entity encoding only includes the location and bounding box of each
object, therefore games with large concave objects (e.g.\ the maze in MsPacman)
are not suitable without some sort of convex decomposition of the objects.

\begin{table}[t]
	\centering
	\begin{tabular}{ l r r r }
		Environment              & CNN          & SA          & GNN          \\
		\hline
		CandyPoison              & \textbf{358} & 233         & 342          \\
		CandyFireballs           & 528          & 283         & \textbf{580} \\
		DispenserFireballs       & 66           & 190         & \textbf{438} \\
		CandyPoison-large        & 177          & 66          & \textbf{225} \\
		CandyFireballs-large     & 231          & 146         & \textbf{235} \\
		DispenserFireballs-large & -6           & \textbf{-4} & -8           \\
		\hline
		Pong (@5.5M)             & -10.07       & -10.19      & 6.57         \\
	\end{tabular}
	\caption{Mean reward acquired by the agent across the episodes. The SPG
		episodes terminate after a fixed 1000 timesteps, while the Atari episodes
		terminate once the agent dies in the game. For Pong we report the result
		after 5.5M environment steps, while for SPG it's 4M, except for the
		$DispenserFireballs$ scenarios where it is 8M.}\label{tab:results}
\end{table}

\begin{figure}[h]
	\centering
		\includegraphics[width=0.4\textwidth]{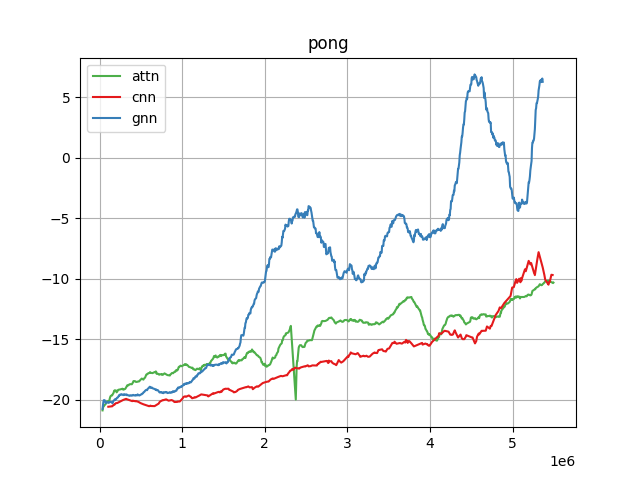}
		\caption{Training curves for Pong using the different architectures. The
			number of iterations is capped at 5M to highlight the benefit of the GNN
			model in the early stages of the traning.}\label{fig:pong_training}
\end{figure}

\section{Conclusion}

We provided an end-to-end approach to RL problems that involves dynamic
observations spaces with unknown number of entities. Our method can be applied
to any sequential decision-making problems that can be captured as a set of
entities that the agent interacts with. We demonstrate the validity of the
approach on the ALE benchmark and on Simple Playgrounds.

We also provide a direct comparison with purely visual RL architectures, and
show that even with a very simple entity-extraction algorithm the training speed
can be increased dramatically, and in complex scenarios that requires multi-step
planning our approach still performs well while the visual method fails to learn
the optimal behaviour.

Furthermore, since our approach only focuses on the relevant bits of
information, instead of processing the whole visual input, it can achieve better
performance with significantly lower number or parameters. In future works a
rigorous comparison will be done to find the lower bound of our approach that
can still find the optimal behaviour in an environment.

Finally, the use of neural networks to learn entity-based representations from
raw sensory inputs is out of the scope of this paper. However it represents a
very promising area for further explorations and experiments. We plan to replace
the simple segmentation module by a learnt segmentation (or object detection)
architecture (either pre-trained or end-to-end), and explore more complex
real-world problems.

\bibliographystyle{named}
\bibliography{ref}

\end{document}